\documentclass{Interspeech}

\interspeechcameraready

\title{Causal Structure Discovery for Error Diagnostics of Children's ASR}

\author[affiliation={1}]{Vishwanath Pratap}{Singh}
\author[affiliation={2}]{Md.}{Sahidullah}
\author[affiliation={1}]{Tomi}{Kinnunen}

\affiliation{School of Computing}{University of Eastern Finland}{Finland}
\affiliation{Institute for Advancing Intelligence}{TCG CREST}{India}
\email{first@university.edu, second@companyA.com, third@companyB.ai}
\email{vsingh@uef.fi, sahidullahmd@gmail.com, tomi.kinnunen@uef.fi}

\usepackage{comment}
\usepackage{multirow}

\begin{document}

\maketitle

\begin{abstract}
Children’s automatic speech recognition (ASR) often underperforms compared to that of adults due to a confluence of interdependent factors: physiological (e.g., smaller vocal tracts), cognitive (e.g., underdeveloped pronunciation), and extrinsic (e.g., vocabulary limitations, background noise). Existing analysis methods examine the impact of these factors in isolation, neglecting interdependencies—such as age affecting ASR accuracy both directly and indirectly via pronunciation skills. In this paper, we introduce a causal structure discovery to unravel these interdependent relationships among physiology, cognition, extrinsic factors, and ASR errors. Then, we employ causal quantification to measure each factor’s impact on children’s ASR. We extend the analysis to fine-tuned models to identify which factors are mitigated by fine-tuning and which remain largely unaffected. Experiments on Whisper and Wav2Vec2.0 demonstrate the generalizability of our findings across different ASR systems. 
\end{abstract}

\section{Introduction}
Automatic speech recognition (ASR) 
is in the growing demand for child-centric technological solutions~\cite{child-tech,child_benchmark,childaugment}. 
Unfortunately, the performance 
of ASR for children lags considerably behind 
that for adults~\cite{child_patel,child-e2e}. \emph{Speech foundation models} (SFM) followed by fine-tuning is the current approach for improving performance for children's ASR~\cite{child_benchmark,jain_whisper, our-csl}. Even after fine-tuning on children's speech, SFMs for children consistently produce higher \emph{word error rates} (WERs)—the standard measure of transcription accuracy—compared to models optimized for adults~\cite{jain_w2v,jain_whisper,kids_hubert,child_benchmark}. This disparity highlights the need for a \textbf{systematic framework} to identify the root causes of accuracy degradation in children's ASR. 

Previous research has identified several factors that contribute to degraded ASR performance for children. Anatomical differences, such as shorter vocal tracts and lighter vocal cords, lead to higher fundamental frequencies and greater spectral variability, which vary with age and gender~\cite{ref5,ref6}. Developing pronunciation skills further contribute to inconsistencies in articulation~\cite{analysis_icslp,impact_of_pronounciation_on_casr_jasa}. 
Children 
may additionally struggle in 
pronouncing complex words despite mastering simpler vocabulary first~\cite{infant_vocab,our-csl}. 
Additionally, in educational settings (where current children's speech data is collected), background babble from parental conversations or classroom sounds tends to interfere with recognition~\cite{myst,cslu}. The length of utterance also plays a role: shorter, fragmented speech tends to be more challenging~\cite{gurunath_csl,our-csl} for attention-based ASR systems~\cite{attention-asr}. 

The above factors are concurrently present in children's speech recordings and hardly impact ASR performance in isolation. Instead, they exhibit complex interdependencies. For instance, age influences both anatomical development and pronunciation ability. Still, 
most 
earlier studies~\cite{analysis_icslp,gurunath_csl,analysis_small_age_groups} 
focus on these factors independently, underscoring a gap in current analysis approaches. Hence, a systematic analysis framework for understanding the causes of degradation should be inclusive of all these causes and should consider their interdependence.

Given that multiple co-existing factors contribute to children's ASR performance 
~\cite{workforce_children,cognitive_linguistic}, a \textbf{causal framework}~\cite{pearl,causal-learn,causal-quantification} provides a natural 
approach for analyzing their interdependencies. Unlike purely \emph{statistical}\footnote{Following Pearl~\cite{pearl}, it is important to be clear on the distinction of correlational ('statistical') and 'causal' terminologies. The former focuses only on statistical associations (such as correlations) and \emph{cannot differentiate between 'cause' and 'effect'}. Causal approaches 'go beyond' correlations and conditional probabilities by encoding explicitly the underlying causal mechanisms, possibly derived from the infamous 'domain knowledge'.} methods that are limited in capturing correlations, a \emph{causal} framework explicitly models cause-and-effect relationships of the underlying mechanisms affecting ASR performance.  In practical terms, causal models are typically formalized using a directed acyclic graph (DAG)~\cite{causal-quantification}. 
The variables under consideration are represented as \emph{nodes} whereas \emph{directed edges} indicate their causal relations. 
Specifically, 
a directed edge from node $A$ to node $B$ in a causal DAG means that $A$ is 
a potential cause of $B$ ~\cite{causal-survey,pearl}. 

For children's ASR analysis, addressed in our work, a causal graph may include nodes for variables such as age, pronunciation skills, background noise, and WER, with the directed edges illustrating how these factors 
influence one another. Using the example of age directly influencing a child's pronunciation skill, and the pronunciation skills in turn impacting WER, a causal DAG may reflect this assumption through a causal chain. 
The DAG helps distinguish \textbf{direct effects} (e.g., age $\rightarrow$ pronunciation) from \textbf{indirect effects} (e.g., age $\rightarrow$ pronunciation $\rightarrow$ WER), enabling a more precise analysis of the 
contributors to ASR performance. 
For the reader less familiar with causal methodology, we provide further detail in
Section~\ref{causal}.


While causality in ASR has been previously studied for different tasks, such as to understand the impact of noise mitigation algorithms on ASR~\cite{asr-error-noise}, the closest related work to ours is~\cite{our-csl}. The authors in that study~\cite{our-csl} used 
a predefined causal DAG to present 
prior knowledge motivated from psychological and social experiments~\cite{workforce_children,cognitive_linguistic}.
Their \emph{hand-crafted} DAG 
hard-codes the assumed 
causal relations between the explanatory variables included. In a stark contrast, in this study we take a much less restrictive approach, 
facilitated by \textbf{causal structure discovery}~\cite{causal-learn}---methodology for identifying 
causal relations \emph{automatically}. 
This considerably simplifies the task of the 
`ASR performance analyst' 
who now only needs to decide which variables (measurements and/or metadata) to include 
to analysis---but without need to specify (potentially restrictive) assumptions on their 
cause-effect relations. 
We argue that reliable and automated identification of the important factors that influence ASR performance is helpful in designing better ASR systems: an ASR engineer benefits from knowing where to focus when designing new architectures, data augmentation or training recipes. Currently we primarily observe only the final outcome effect---high WERs in children ASRs---but lack accurate picture of the root causes. 


\begin{table}[t]
    \centering
    \caption{Categorization of variables in analysis. Variables not available in metadata of CSLU Kids~\cite{cslu} are inferred, and their computation is discussed in Section~\ref{sec:exp}.}
    \vspace{-0.1cm}
    \begin{tabular}{|c|c|c|}
        \hline
        \textbf{Type} &\textbf{Variable} & \textbf{Metadata/Inferred} \\ 
        \hline
        \multirow{2}{*}{Physiological} & Age & Available  \\
        \cline{2-3}
                                       & Gender & Available \\
        \hline
        Cognitive & Pronunciation Ability & Inferred  \\
        \hline
        \multirow{3}{*}{Extrinsic} & Signal-to-Noise Ratio & Inferred \\
        \cline{2-3}
                                   & Vocab Difficulty & Inferred \\
        \cline{2-3}
                                   & \#Words in Audio & Available \\
        \hline
    \end{tabular}
    \label{tab:table-var}
   
\end{table}

We use the same set of variables and categorization  (see Table~\ref{tab:table-var}) as~\cite{our-csl} for easier comparability with that prior work. The ASR systems included consists of two well-known categories of open-source SFMs, representative of present state-of-the-art: (i) self-supervised -- \texttt{Wav2Vec2.0}~\cite{wav2vec}, and (ii) weakly-supervised -- \texttt{Whisper}~\cite{whisper}. 
We consider both off-the-shelf (pretrained) and 
fine-tuned SFMs. We utilized CSLU kids corpus~\cite{cslu} for causal structure discovery, as it has gender and age metadata available from a diverse range. While we utilize MyST~\cite{myst} for fine-tuning the SFMs.  

\begin{table*}[h!]
\centering
\vspace{-0.2cm}
\caption{Summary of Causal methods being used for explainable machine learning, and comparison with our proposed approach.}
\begin{tabular}{|c|c|c|c|}
\hline
\textbf{Work} & \textbf{Task} & \textbf{Causal Structure Discovery Method} & \textbf{Causal Quantification Method} \\
\hline
\cite{animal} & Animal Behavioral Modeling  & PC-MI~\cite{PCMI} & Graph Neural Network \\ 
\hline
\cite{ci_reco} & Explainability in Recommendation System& Domain Knowledge & Logistic Regression \\
\hline
\cite{zhou2024} & Sentiment Classification in NLP & Domain Knowledge & Bayesian Network\\
\hline

\cite{our-csl} & Causes of degradation in children's ASR & Prior Knowledge & Bayesian Network \\
\hline
Our & Causes of degradation in children's ASR & PC~\cite{pc} and FCI~\cite{fci} & Bayesian Network \\
\hline
\end{tabular}
\vspace{-0.2cm}
\label{tab:ci}
\end{table*}
\section{A Primer in Causality}
\label{causal}
Causal analysis aims to establish cause-and-effect relationships that go beyond mere statistical (correlational) associations~\cite{causal-survey, causal-quantification}. 
Causal relations are formalized through a \emph{directed acyclic graph} (DAG), whose nodes 
represent the variables 
and edges represent the cause-effect relations. 
 Causal analysis is typically conducted in two stages: (1) Causal structure discovery, which involves identifying causal relationships among variables in a DAG—either hardcoded from prior knowledge~\cite{our-csl,ci_reco} or learned from data~\cite{animal}; and (2) Causal quantification, which focuses on estimating the functional relationships between the connected variables. We review each step briefly.

\subsection{Causal Graph}
\begin{figure}[!t]
\vspace{-0.2 cm}
    \centering
    \begin{minipage}{0.08\textwidth}
        \centering
        \includegraphics[width=\textwidth]{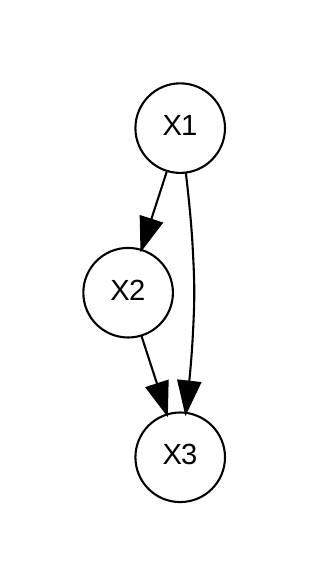}
        (a)
        \label{fig:confounder}
    \end{minipage}%
    \hfill
    \hspace{0.5 cm}
    \begin{minipage}{0.068\textwidth}
        \centering
        \includegraphics[width=\textwidth]{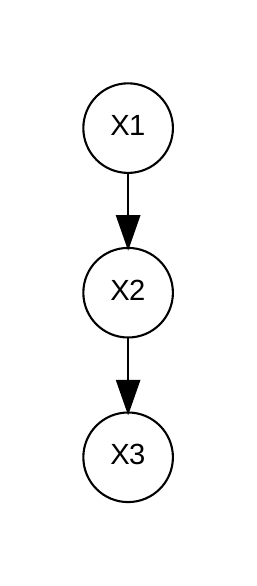}
        \label{fig:chain}
        (b)
    \end{minipage}
    \hfill
    \begin{minipage}{0.14\textwidth}
        \centering
        \includegraphics[width=\textwidth]{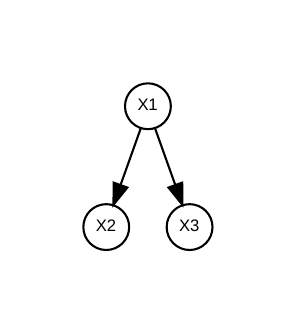}
        \label{fig:fork}
        (c)
    \end{minipage}
    \hfill
    \begin{minipage}{0.14\textwidth}
        \centering
        \includegraphics[width=\textwidth]{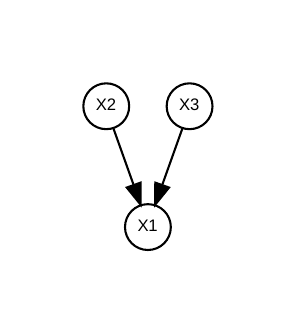}
        (d)
        \label{fig:collider}
    \end{minipage}
    \caption{Examples of directed acyclic graphs (DAGs) illustrating the causal relationships. (a.) Confounder: where $X_2$ is a parent of $X_3$, $X_1$ is a parent of both $X_2$ and $X_3$, making $X_1$ a confounder. $X_2$ and $X_3$ are descendants of $X_1$. (b.) Chain:  a node that lies sequentially between two other nodes. (c.) Fork: indicates the nodes that have multiple children. (d.) Collider: a node is influenced by two or more nodes. }
    \label{dag}
    \vspace{-0.0cm}
\end{figure}

A causal graph, represented using DAG, with \( N \) random variables encodes how interventions on one affect others. It has generally four node types (Fig.~\ref{dag}). Formally, let \( G \) be a causal DAG with nodes \( X_1, \ldots, X_n \).
An edge $X_i \rightarrow X_j$ implies that $X_i$ causally influences $X_j$. In the example shown in Fig.~\ref{dag}(a), $X_1$ has no parents, $X_2$ has $X_1$ as its parent (denoted by $\text{PA}_2 = \{X_1\}$), $X_3$ has $X_2, X_1$ as its parents ($\text{PA}_3 = \{X_2, X_1\}$), and $X_2$ and $X_3$ are descendants of $X_1$. 


\subsection{Causal Structure Discovery}
Traditional methods for uncovering causal relationships rely on 
\emph{interventions} or \emph{randomized experiments} (studies where variables are deliberately manipulated to observe causal effects, such as randomized controlled trials, which randomly assign subjects to different groups). They can be impractical in machine learning due to cost and feasibility constraints—such as the need for extensive resources during data collection, and ethical considerations~\cite{causal-learn}. 
Causal structure discovery achieves this by inferring causal relationships from observational data—data collected \emph{without} direct intervention, simply by observing variables as they occur naturally~\cite{causal-learn2}. In this context (i.e., inferring from observational data), causal structure discovery serves as a \emph{post-hoc} explainability approach, providing explanations for decisions \emph{after} they have been made~\cite{kim2016, renkl2014}.

A naive approach to causal structure discovery would involve evaluating all possible edge configurations among \( N \) variables. Since each directed edge can either be present or absent, the total number of possible configurations scales as \( \mathcal{O}(2^{N}) \), which is exponential in \( N \). This 
makes brute-force exploration infeasible, even for relatively small \( N \). Consequently, more efficient algorithms are necessary to infer causal structures in a computationally tractable manner~\cite{pc}.

Two commonly used causal structure discovery algorithms include \emph{Peter-Clark} (PC)~\cite{pc} and \emph{Fast Causal Inference} (FCI)~\cite{fci} algorithms. Both 
leverage conditional independence~\cite{conditional-independence} tests to extract information about the underlying causal structure. The main difference between the two is that 
PC assumes so-called \emph{causal sufficiency} (i.e., no hidden confounders), whereas FCI accounts for latent variables and selection bias, making it more suitable for scenarios with unmeasured confounders.

\subsection{Causal Quantification}
Once the causal relationship between variables is established (i.e., DAG is formed), the next step is to estimate the functional relationship between the nodes, a process known as \emph{causal quantification}~\cite{causal-quantification}. 
Whereas a directed edge (whether hand-crafted or automatically discovered) from age to WER in a DAG signifies that age has a causal effect on WER, causal quantification further determines the \emph{magnitude} of this effect--e.g., 
how much 
an increase in age by one year increases (or decreases) WER. 

A common approach to quantifying causal strength in DAGs is the \emph{average causal effect} (ACE)~\cite{causal-quantification}, which measures the expected outcome difference between two scenarios when a node in DAG takes different values (e.g., 0 vs. 1). ACE is defined as:
\begin{equation}
    \text{ACE}(X_i \rightarrow X_j) = \mathbb{E}[X_j \mid X_i = x_1] - \mathbb{E}[X_j \mid X_i = x_0]
\end{equation}


This allows us to quantify the causal impact of specific variables on ASR performance, helping identify which factors considerably influence outcomes such as word error rates.

\subsection{Causality in Machine Learning}
Causality-aided explainable machine learning has gained popularity recently. A prior study~\cite{our-csl} provides an overview of causality in ML; we briefly compare these studies with our approach in Table~\ref{tab:ci}.

For direct comparison with prior work~\cite{our-csl}, we use \emph{Bayesian inference} for causal quantification. First, we perform causal structure discovery and quantification to assess how open-source SFMs respond to variations in children's physiology (age, gender~\cite{workforce_children}), cognition (e.g., pronunciation~\cite{cognitive_linguistic}), and extrinsic factors. Then, we repeat the evaluation with fine-tuned SFMs.

\section{Experimental Setup}
\label{sec:exp}
\subsection{Dataset}  
Following~\cite{our-csl}, we use CSLU Kids~\cite{cslu} for causal analysis due to the availability of diverse age groups and gender metadata. Since no standard protocol (training/development/test split) is available for CLSU Kids, 
we use the 
publicly available protocol from~\cite{childaugment,our-csl}. 
We also include MyST~\cite{myst} dataset for fine-tuning the SFMs to enhance generality of findings reported. 
For MyST, a standard 
protocol is available. 
Different from CSLU Kids, unfortunately, age and gender metadata labels are not available. 
This limits the use of the otherwise more recent and larger MyST in causal analysis. 

\subsection{Automatic Speech Recognition Systems}
We consider two speech foundation model (SFM) based ASR systems in our experiments: \texttt{Wav2Vec 2.0} and \texttt{Whisper}.  These two models complete the spectrum of SFMs, representing the two broad categories of self-supervised and weakly-supervised approaches. We utilize the open-source pre-trained \texttt{Whisper-Small}~\footnote{https://huggingface.co/openai/whisper-small} 
and \texttt{Wav2Vec2.0-large}~\footnote{https://huggingface.co/facebook/wav2vec2-large-960h-lv60-self} 
for evaluation and fine-tuning.  
First, we use open-source SFMs "as is" in analysis then we fine-tune on MyST dataset~\cite{myst}. 

\subsection{Inferred Variables}
In this section, we describe how inferred variables in Table~\ref{tab:table-var} are computed. For \emph{pronunciation ability}, we use the Goodness of Pronunciation (GoP) score~\cite{gop}, a posterior probability of the tartget phone normalized by 
the maximum posterior of all phones.
Similar to~\cite{our-csl}, per-phoneme scores are averaged to produce an utterance-level score, which is further discretized into \emph{Low}, \emph{Average}, and \emph{High} values. 
For \emph{Vocabulary Difficulty}, we employ a rarity-based metric~\cite{word_difficulty} using word frequencies extracted from multiple 
text corpora. 
A sentence-level difficulty score is obtained by averaging individual word scores and discretizing the result into \emph{Low}, \emph{Average}, and \emph{High} levels. Finally, for signal-to-noise (SNR) estimation, we use 
NIST's 
toolkit~\footnote{https://www.nist.gov/information-technologylaboratory/iad/mig/nist-speech-signal-noise-ratio-measurements}. Similar to GoP and Vocabulary Difficulty, we discretize SNR to three categories: \emph{Clean} (SNR $\geq$ 20 dB), \emph{Average} ($5 \leq \text{SNR} \leq 20$ dB), and \emph{Noisy} (SNR $\leq$ 5 dB). Discretization allows us to stratify the GoP, Vocabulary Difficulty, and SNR into meaningful groups, which helps in identifying cause-effect relationships more precisely~\cite{bishop}. 

\section{Results and Discussion}
Unlike \cite{our-csl}, which uses a hardcoded DAG and focuses solely on causal quantification, we perform both causal structure discovery and causal quantification.
\begin{figure*}[!t]
\vspace{-0.0cm}
    \centering
    \begin{minipage}{0.4\textwidth}
        \centering
        \includegraphics[width=\textwidth]{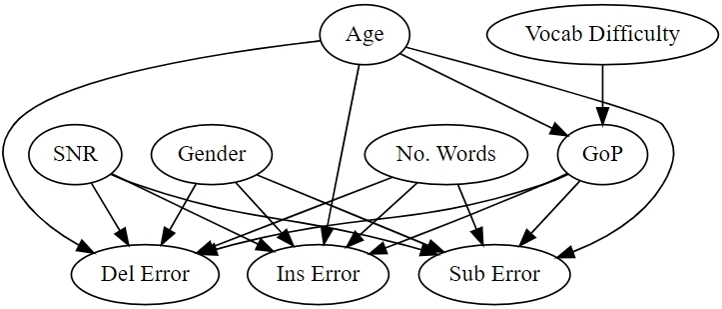}
        \label{fig:prior}
        {(a)}
    \end{minipage}%
    \hfill
    \begin{minipage}{0.35\textwidth}
        \centering
        \includegraphics[width=\textwidth]{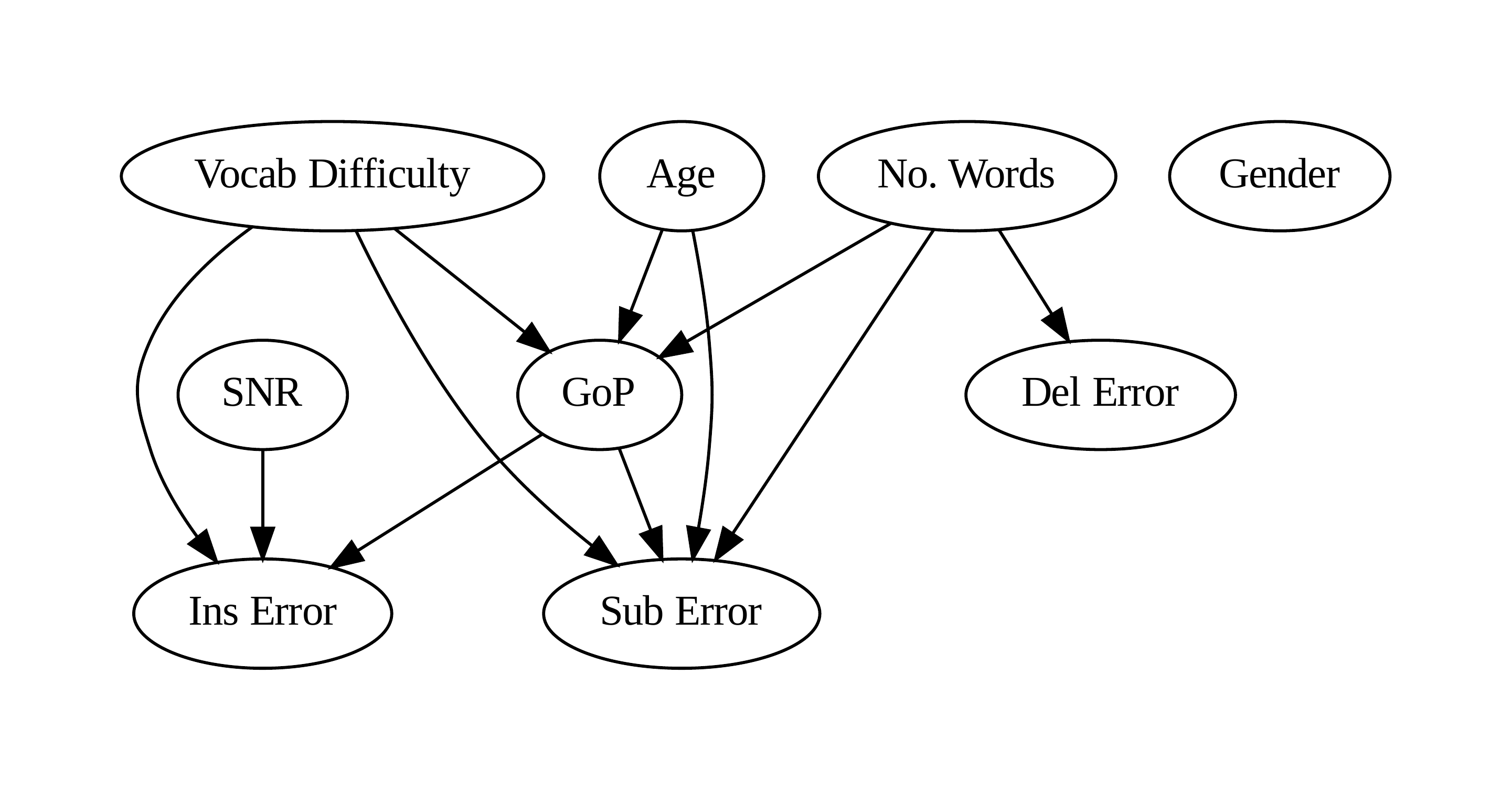}
        \label{fig:pc}
        {(b)}
    \end{minipage}
    \vspace{-0.0cm}
    \caption{Causal structure framework for explaining the ASR errors for children. (a.) Hard-coded: directed acyclic graph (DAG) used in prior study~\cite{our-csl}. (b.) Data-driven: Causal structure discovery-based DAG using PC and FCI algorithms. Both PC and FCI have produced the same DAG. }
    \label{fig:dag}
    \vspace{-0.0cm}
\end{figure*}

\begin{table*}[htb]
\renewcommand{\arraystretch}{1.1}
\centering
\caption{Comparison between the average causal effect (ACE) for hardcoded DAG vs data-driven DAG for open-source and fine-tuned \texttt{Whisper-small} and \texttt{Wav2Vec2.0-large} models. Empty cells corresponds to absent edges in data-driven DAG.}
\vspace{0cm}
\label{tab:res1}
\begin{tabular}{|c|c|c|c|c|c|c|c|c|c|c|c|c|c|}
\hline
\multirow{3}{*}{\textbf{Model}} & \multirow{3}{*}{\textbf{Cause}} & \multicolumn{6}{c|}{\textbf{Open-source}} & \multicolumn{6}{c|}{\textbf{Fine-tuned}} \\
\cline{3-14}
 & & \multicolumn{3}{c|}{\textbf{Hardcoded DAG}} & \multicolumn{3}{c|}{\textbf{Data driven DAG}} & \multicolumn{3}{c|}{\textbf{Hardcoded DAG}} & \multicolumn{3}{c|}{\textbf{Data driven DAG}} \\
\cline{3-14}
 & & \textbf{Subs} & \textbf{Del} & \textbf{Ins} & \textbf{Subs} & \textbf{Del} & \textbf{Ins} & \textbf{Subs} & \textbf{Del} & \textbf{Ins} & \textbf{Subs} & \textbf{Del} & \textbf{Ins} \\
\hline
\multirow{5}{*}{\texttt{Wav2Vec2.0}} & Age & -4.36 & -0.10 & -1.60 & -5.21 & -- & -- & -2.12 & -0.04 & -2.81 & -3.58 & -- & -- \\
\cline{2-14}
 & Gender & 0.80 & 0.17 & 0.87 & -- & -- & -- & 0.25 & 0.06 & 0.22 & -- & -- & -- \\
\cline{2-14}
 & GoP & -1.07 & -0.25 & -0.88 & -1.27 & -- & -1.08 & -0.63 & -0.14 & -0.54 & -0.84 & -- & -0.45 \\
\cline{2-14}
 & SNR & -1.08 & 0.09 & 0.28 & -- & -- & -2.38 & -0.58 & 0.11 & 0.09 & -- & -- & -1.75 \\
\cline{2-14}
 & \# Words & -9.20 & 0.30 & -4.23 & -12.1 & 0.25 & -- & -7.14 & 0.36 & -3.12 & -9.74 & 0.08 & -- \\
\hline
\multirow{5}{*}{\texttt{Whisper}} & Age & -3.25 & -0.20 & -2.16 & -3.97 & -- & -- & -3.0 & 0.32 & -2.22 & -2.31 & -- & -- \\
\cline{2-14}
 & Gender & 0.62 & 0.37 & 1.15 & -- & -- & -- & 1.90 & 0.10 & 1.33 & -- & -- & -- \\
\cline{2-14}
 & GoP & -1.25 & -0.18 & 0.35 & -0.9 & -- & -0.22 & -1.10 & -0.49 & -0.05 & -0.65 & -- & -0.18 \\
\cline{2-14}
 & SNR & -1.21 & 0.14 & -0.21 & -- & -- & -1.84 & -1.02 & 0.15 & -0.36 & -- & -- & -1.54 \\
\cline{2-14}
 & \# Words & -5.10 & 1.02 & -3.45 & -4.76 & 0.85 & -- & -5.53 & 1.20 & -3.59 & -4.80 & 0.95 & -- \\
\hline
\end{tabular}
\vspace{-0.0cm}
\end{table*}

\subsection{Causal Structure Discovery}
First, we examine the differences between the hardcoded DAG used in a prior study~\cite{our-csl} and the data-driven DAG obtained using PC and FCI, as shown in Figure~\ref{fig:dag}. Below, we discuss the similarities and differences for each node:

Similar to the hardcoded DAG, the 
automatically inferred DAG suggests that physiological factors, such as age, influences both cognitive factors (e.g., pronunciation variability) and ASR errors. However, unlike the hard-coded DAG, which assumes that age affects all three types of ASR errors, the data-driven DAG indicates that age variability primarily contributes to substitution errors.

Regarding 
gender, the hardcoded DAG assumes it to 
influences ASR errors, whereas the data-driven DAG indicates no causal relationship between gender and ASR errors. This aligns with previous empirical studies~\cite{our-csl,child-e2e}, which found no considerable differences in ASR errors between boys and girls speakers.

For pronunciation variation (GoP), both DAGs identify Vocabulary Difficulty and Age as influencing pronunciation variability. However, the data-driven DAG also incorporates an additional factor—the number of spoken words (i.e., sentence length). Moreover, the data-driven DAG suggests that mispronunciations predominantly lead to substitution and insertion errors.

Regarding SNR, the hardcoded DAG assumes that SNR affects all three types of ASR errors, whereas the data-driven DAG suggests that SNR (representing babble noise in a classroom setting) primarily contributes to insertion errors. This aligns with intuition, as overlapping background speech can introduce unintended words into ASR transcriptions, leading to insertion errors.

Lastly, concerning utterance length, the hardcoded DAG assumes that all three types of ASR errors are influenced by the number of words in an audio sample. In contrast, the data-driven DAG suggests that content length primarily affects substitution and deletion errors.

\subsection{Causal Quantification}

In this section, we compare the analysis results obtained using these two different frameworks. Quantification of ASR errors using ACE for children is presented in Table~\ref{tab:res1}. First, for open-source SFMs, we observe that similar to the hardcoded DAG, the ACE for Age is negative for both \texttt{Wav2Vec2.0} and \texttt{Whisper} models in the data-driven DAG, indicating that an increase in Age reduces ASR errors. However, the hardcoded DAG shows a larger absolute ACE for substitution errors and a smaller ACE for deletion and insertion errors. This suggests that Age plays a lesser role in these two types of errors—interestingly, these are the very errors for which the data-driven DAG does not have a direct edge from Age.

Similarly, in the hardcoded DAG, the ACE for deletion and insertion errors is smaller for the \emph{number of words in audio} (No. of Words), while the data-driven DAG lacks a direct edge between No. of Words and these two error types. Similar observations can be made regarding the lower ACE for Gender in hardcoded DAG while absent edges for Gender in data-driven DAG. Hence, forced causal associations in hardcoded DAG~\cite{our-csl} result into weaker causal relationships (in terms of ACE) while these weaker edges are not present in the data-driven DAG. 

Finally, we present the ACE for both hardcoded and data-driven DAGs for fine-tuned \texttt{Whisper} and \texttt{Wav2Vec2.0} models in Table~\ref{tab:res1}. Our findings indicate that similar to hardcoded DAG, data-driven DAG also shows a reduction in ACE for for fine-tuned model. Specifically, after fine-tuning both DAGs have shown considerably lower ACE for Age (than that of open-source model in Table ~\ref{tab:res1}), indicating that fine-tuning reduces the impact of Age on ASR errors. However, ACE for \emph{No. Words} node in both DAGs remain very high, even after fine-tuning indicating the limitation of fine-tuning.

\section{Conclusion}
We presented an approach for the construction of causal graphs for analyzing ASR errors in children. Unlike prior studies with hardcoded causal link assumptions, our data-driven method learns the causality relations automatically and removes unnecessary edges from the causal graph, thereby simplifying the analysis. 

ACE analysis identifies key factors impacting ASR performance, guiding future research on data selection, model adaptation, and preprocessing for improved accuracy. For instance, fine-tuning addresses the age factor in ASR errors, while ACE for shorter utterances (which are typically in interactions of children's digital devices) remains very high. Hence, future research can be oriented towards addressing this concern by including short utterances in training or architecture suitable for short utterances.

One limitation of our study is its reliance on the CSLU Kids dataset due to the scarcity of child speech datasets with necessary metadata. Future work will explore additional datasets and extend causal inference to broader speech-processing tasks.



\newpage
\ifinterspeechfinal
\section{Acknowledgements}
The authors wish to acknowledge CSC – IT Center for Science, Finland, for computational resources.
\bibliographystyle{IEEEtran}
\bibliography{mybib}

\else
\bibliographystyle{IEEEtran}
\bibliography{mybib}
\fi

\end{document}